\documentclass[conference]{IEEEtran}
\IEEEoverridecommandlockouts

\usepackage{cite}
\usepackage{amsmath,amssymb,amsfonts}
\usepackage{algorithmic}
\usepackage{graphicx}
\usepackage{textcomp}
\usepackage{xcolor}
\def\BibTeX{{\rm B\kern-.05em{\sc i\kern-.025em b}\kern-.08em
    T\kern-.1667em\lower.7ex\hbox{E}\kern-.125emX}}

\usepackage{multicol, multirow, booktabs}
\usepackage{colortbl}
\usepackage{mathtools}
\usepackage[linesnumbered,ruled,vlined]{algorithm2e}
\SetKwInput{Input}{Input}
\SetKwProg{kwServer}{\textcolor{blue}{Server Update}}{}{}
\SetKwProg{kwClient}{\textcolor{blue}{Client Update}}{}{}
\SetKwProg{init}{\textcolor{blue}{Initialization}}{}{}
\SetKwInOut{Output}{Output}
\SetKwProg{procedure}{procedure}{}{}

\begin{document}

\title{Securing Federated Learning against Backdoor Threats with Foundation Model Integration}

\author{\IEEEauthorblockN{1\textsuperscript{st} Xiaohuan Bi}
\IEEEauthorblockA{
\textit{Renmin University of China}\\
2021201555@ruc.edu.cn}
\and
\IEEEauthorblockN{2\textsuperscript{nd} Xi Li}
\IEEEauthorblockA{
\textit{The University of Alabama at Birmingham}\\
XiLiUAB@uab.edu}
}

\maketitle

\begin{abstract}
Federated Learning (FL) enables decentralized model training while preserving privacy. Recently, the integration of Foundation Models (FMs) into FL has enhanced performance but introduced a novel backdoor attack mechanism. 
Attackers can exploit FM vulnerabilities to embed backdoors into synthetic data generated by FMs. During global model fusion, these backdoors are transferred to the global model through compromised synthetic data, subsequently infecting \textit{all} client models.
Existing FL backdoor defenses are ineffective against this novel attack due to its fundamentally different mechanism compared to classic ones. 
In this work, we propose a novel data-free defense strategy that addresses both classic and novel backdoor attacks in FL. 
The shared attack pattern lies in the abnormal activations within the hidden feature space during model aggregation.
Hence, we propose to constrain internal activations to remain within reasonable ranges, effectively mitigating attacks while preserving model functionality. 
The activation constraints are optimized using synthetic data alongside FL training. 
Extensive experiments demonstrate its effectiveness against both novel and classic backdoor attacks, outperforming existing defenses.
\end{abstract}

\begin{IEEEkeywords}
Backdoor Defense, Federated Learning, Foundation Models, Adversarial Machine Learning
\end{IEEEkeywords}

\section{Introduction}\label{sec:intro}


Federated learning (FL) enables models to be collaboratively trained across multiple decentralized institutes or devices. 
This method significantly enhances data privacy and security and has a wide range of applications, including healthcare\cite{healthcare,wang2022}, finance\cite{finance}, and IoT\cite{iot}.
Recently, the integration of Foundation Models (FMs) like the GPT series \cite{gpt}, LLaMA \cite{llama}, and Stable Diffusion \cite{stable_diffusion} has introduced new dynamics to FL\cite{FMinFL}. 
These models, pre-trained on vast and varied datasets, show remarkable proficiency in various tasks from natural language processing to vision data generation and recognition. 
By leveraging their extensive pre-trained knowledge, FMs can augment FL systems either by enhancing performance through knowledge distillation \cite{FedMD,abs-2303-10917} 
or by facilitating client knowledge sharing with synthetic data generation 
\cite{FedDF,assefa2020generating}.

However, incorporating FMs into FL systems (FM-FL) introduces new complexities and exacerbates existing threats. 
Recent studies \cite{BD_FMFL,BD_FMHFL,BD_FMFL_all,FMFL_reliable} have identified a novel backdoor mechanism in FL arising from the interaction between FL and FM. 
Consider a classic FM-FL scenario where the FM on the server generates synthetic data that mirrors the client data distribution for use in global model aggregation. 
An attacker can exploit FM vulnerabilities to embed a backdoor into the synthetic data.
This backdoor is then planted into the global model during the aggregation stage using the poisoned synthetic dataset.
Subsequently, the backdoor is transmitted to \textit{all} FL clients as the poisoned global model is distributed, thereby compromising the integrity of the entire system.

Existing FL backdoor defense methods cannot address the novel attack due to its fundamentally different attack mechanism compared to the classic one.
In the classic attack, a few clients are compromised first, and the global model is subsequently infected through client model fusion. 
To mitigate such attacks, existing methods detect \textit{anomalies} in client updates and exclude them from model aggregation \cite{NormThr, Krum, clipcluster, signguard, rfout}.
In contrast, the novel attack introduced by the integration of FMs compromises the global model first through the use of poisoned synthetic data, which then propagates the backdoor to all client models. 
Since the local data remains untouched, all client updates appear similar, and no detectable anomalies arise. 
As a result, existing FL defenses are ineffective against this type of attack.
Moreover, many general machine learning (ML) backdoor defenses are unsuitable. 
Some require access to clean, real datasets \cite{NAD,Fine-Pruning}, conflicting with FL's data isolation principle, while others are time-intensive \cite{NC}, making them impractical for continuous FL training cycles.

It is critical to develop an effective defense strategy that addresses both novel and classic backdoor threats within the FL framework.
A common attack pattern in both cases is the abnormally large activations in hidden layers of compromised models when processing manipulated inputs \cite{MMBD, Fine-Pruning}.
To mitigate this, a feasible approach is to impose upper bounds on internal activations.
These bounds constrain activations to reasonable limits, effectively mitigating malicious behavior on poisoned samples while preserving model functionality on normal ones.
The optimization of these bounds leverages synthetic data, maintaining local data independence in FL.
This process is applied after global model fusion on the server, with original model parameters fixed, ensuring both functionality and defense efficiency.
Hence, this method provides effective defense against backdoor attacks while remaining well-suited to the FL framework.

In summary, this paper presents the following contributions:
(1) We propose the first data-free defense strategy that addresses both novel and classic backdoor attacks in FL with FM integration.
(2) Extensive experiments across diverse FL scenarios validate the effectiveness of our unified defense framework against both types of backdoor attacks.

\section{Related Work}\label{sec:rw}

\subsection{Backdoor Attacks in FL}
A backdoor attacker in FL aims to embed malicious behavior into the global model distributed to all clients. 
This backdoor behavior (\textit{e.g.}, misclassification to a specific target class) is triggered only by specific patterns embedded in input samples, while the model functions normally on clean inputs.

\textbf{Classic Backdoor Threats:}
Classic backdoor threats primarily target the client side through techniques like data poisoning \cite{tolpegin2020data}, local model poisoning \cite{fang2020local}, and attacks such as semantic and distributed backdoors \cite{BD_FL, DBLP:conf/nips/WangSRVASLP20, DBLP:conf/iclr/XieHCL20}. 
For instance, attackers may inject poisoned samples into the local training datasets of compromised clients. These compromised local models then propagate the malicious model updates to the global model during server-side aggregation. 
With sufficient compromised clients and communication rounds, the global model is embedded with backdoor threats.

\textbf{Novel Backdoor Threats:}
Recent studies \cite{BD_FMFL,BD_FMHFL} have revealed a novel backdoor attack mechanism in FL that exploits the interaction between FMs and FL. 
Attackers leverage vulnerabilities in FMs \cite{ICL_survey,shi2023badgpt,xu2023instructions} to embed backdoors into FM-generated synthetic data. This poisoned data is subsequently used for purposes such as model pre-training \cite{GPT-FL}, knowledge transfer \cite{abs-2303-10917}, or model aggregation \cite{FedDF}. 
The backdoor threat is then incorporated into the global model through the use of synthetic data and subsequently transmitted to all FL clients.

\textbf{Classic vs. Novel Attack Mechanisms:}
Classic backdoor threats stem from compromised clients and propagate to the server, with malicious updates identified as anomalies among client updates. In contrast, novel attacks originate on the server, embedding threats into the global model and affecting all clients uniformly, making anomalies undetectable.


\subsection{FL Backdoor Defenses} 
Existing FL defenses \cite{NormThr,DP,Krum,clipcluster,signguard,rfout} primarily target client-originated threats and offer limited protection against novel attack approaches. 
These methods focus on identifying anomalies in client updates and excluding the updates from identified malicious clients during global model aggregation to achieve a robust global model.
Specifically,
\cite{NormThr} applies a fixed norm threshold to client updates in FL.
\cite{DP} combines norm thresholding with the injection of Gaussian noise into the aggregated global model at the server.
\cite{Krum, clipcluster, signguard, rfout} select the most reliable gradient updates from all participants to counter adversarial or faulty updates.
\cite{Pruning} utilizes federated aggregation of neuron activation values to prune the least active neurons.

\section{Methodology}\label{sec:method}


\subsection{Overview}
\textbf{FM integration in FL.}
Our work follows existing FM-integrated FL (FM-FL) frameworks, such as as those proposed in \cite{FMinFL,GPT-FL}. 
The basic FM-FL cycle, as illustrated in Fig.\ref{fig:defense} and Algorithm\ref{alg:defense}, consists of three key steps.
\textbf{Stage 1: Initialization.}
An FM is integrated into the server to generate synthetic data (\textit{e.g.}, text or image data) that mirrors the distribution of client-local data and is later used to fuse a global model.
\textbf{Stage 2: Client Update.}
Clients independently train their local models using private local data. Once trained, they upload their model parameters to the server for aggregation during the model fusion process.
\textbf{Stage 3: Server Global Model Fusion.}
The server aggregates the client model parameters using synthetic data as a carrier for client model information sharing. 
This process employs aggregation functions such as those proposed in \cite{FedDF,FedMD}, which are applicable to various FL settings.

\textbf{The proposed defense.}
To provide a unified defense strategy against both classic and novel backdoor attacks in FL, we propose \textbf{Stage 4: Server Backdoor Mitigation}, implemented on the server after the global model fusion.
Regardless of whether the backdoor attacks are novel or classic, poisoned samples typically exhibit abnormal internal activations.
Hence, the proposed defense optimizes an upper bound on the internal activations of the global model. 
This bound constrains the abnormal activations caused by poisoned samples while allowing the normal activations of clean samples to pass.

Stages 2–4 are repeated until FL converges. Throughout this process, we aim to minimize poisoning impact while maintaining effective training and aggregation.

\subsection{Threat Model and Assumptions on Defender}
Our threat model aligns with the use of cutting-edge FMs accessed via APIs and focuses on classification tasks, which is commonly studied in both backdoor and FL research.

\noindent\textbf{Attacker’s knowledge:} 
The attacker lacks access to the local training set and process, distinguishing it from traditional backdoor attacks. 
Instead, they exploit access to the server’s FM queries to insert backdoor instructions, specifying the trigger, target class, and demonstrations that show how the attack is activated \cite{BD_FMFL,BD_FMHFL}.

\noindent\textbf{Attacker’s goals:} 
The attacker aims to 
(1) guide the FMs to generate synthetic datasets containing backdoor-poisoned samples, and
(2) leveraging (1), propagate the backdoor to all client models in FL, causing the final model to misclassify triggered inputs to the target class while maintaining high performance on clean samples.


\noindent\textbf{Defender’s knowledge:} 
Like most FL defenses, this method is server-side, with no access to clean local training sets or processes. 
It assumes no knowledge of an attack, trigger type, or target class, relying solely on local updates and the server’s synthetic dataset.

\noindent\textbf{Defender’s goals:} 
The defender aims to mitigate backdoor attacks during FL training, ensuring the final model matches clean model performance.
Specifically, it should classify backdoor-embedded inputs correctly while maintaining high accuracy on clean samples.

\subsection{Server-side Backdoor Mitigation}

\begin{figure}
    \centering
    \includegraphics[width=.9\linewidth]{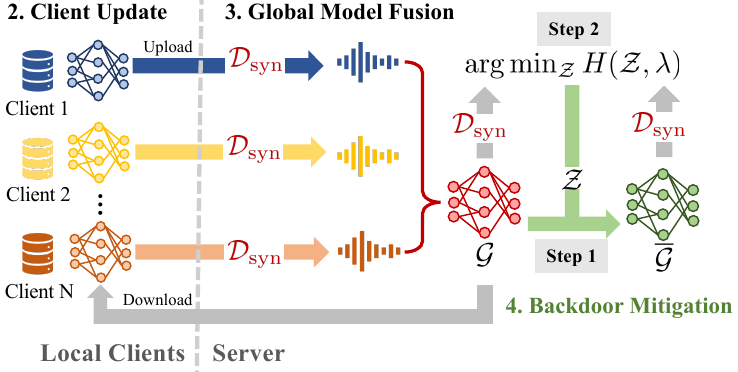}
    \vspace{-0.05in}
    \caption{Our defense strategy for the FM-FL framework.}
    \label{fig:defense}
    \vspace{-0.2in}
\end{figure}

Existing defenses, initially proposed for classic backdoor attacks that arise from compromising several clients, face challenges with the novel backdoor attack.
\textbf{Challenge 1.} 
Most FL defenses \cite{NormThr,Krum,signguard,rfout,clipcluster} detect \textit{anomalies} in clients but are ineffective against novel attacks where malicious updates may come from the \textit{majority} of clients.
\textbf{Challenge 2.} 
General ML defenses \cite{NAD,Fine-Pruning} need clean, real datasets.
This contradicts the decentralized nature of FL, where no real data is available on the server.
\textbf{Challenge 3.} 
The ongoing nature of FL makes repeated use of \textit{time-consuming} ML defenses like trigger estimation \cite{NC} impractical.

\textbf{To address Challenge 1}, we focus on the underlying attack pattern common to both novel and classic backdoor attacks in FL.
Whether the attack originates from a few compromised clients or arises through the interaction between FM and FL, the poisoned global model exhibits the same malicious pattern: abnormally large activations in hidden layers triggered by poisoned inputs.
This pattern has been observed in prior studies \cite{MMBD, Fine-Pruning}.
Leveraging this shared malicious pattern, we propose transitioning from traditional client-level anomaly detection to activation-level anomaly detection, offering a unified perspective on backdoor defense. 
\textit{Malicious updates can be mitigated by imposing upper bounds on internal activations of the global model. }
These bounds are optimized to maintain model performance on specific datasets.
\textbf{To address challenge 2}, upper bounds optimization can be performed using the (possibly poisoned) synthetic dataset, preserving the decentralized nature of FL. 
This method also minimizes impact on the model's classification ability as it does not alter underlying parameters. 
Finally, \textbf{to address Challenge 3}, periodically adjusting these bounds is a more efficient and practical alternative to the extensive parameter tuning or trigger estimation required in other defenses. 

Motivated by the above, we propose imposing an upper bound on the internal activations of the global model to constrain abnormal activations caused by backdoor threats.
The defense is implemented after global model aggregation on the server and consists of three main steps: \textbf{upper-bound application, optimization, and hyper-parameter adjustment}, as illustrated in Fig.\ref{fig:defense} and Algorithm\ref{alg:defense}.
We will elaborate on the server-side defense in FL in the following sections. 
Note that the core FL stages, such as client updates and server-side global model fusion, remain unchanged and are not covered here. For details, please refer to \cite{FedDF}.

\textbf{Step 1: Upper Bound Application.}
Let $\mathcal{G}(\cdot)$ denote the $L$-layer global model, derived from the model fusion at FL round $t$\footnote{For clarity, we omit $t$ in our notation.}. 
We define the set $\mathcal{K} \subseteq \{1,\cdots,L\}$ to include all layers where bounds are applied, and the set $\mathcal{Z} = \{\mathbf{z}_k \in \mathbb{R}^{n_k} | k \in \mathcal{K}\}$ contains the upper bound vectors for activations constrain at layer $k$ within $\mathcal{K}$.
Then the logit function for any class $c\in \mathcal{C}$ and any input $\mathbf{x}\in \mathbb{R}^{n_{0}}$ with activation bounds $\mathcal{Z}$ can be written as
\begin{equation}\label{eq:bounded_activation}
    \overline{\mathcal{G}}_c(\mathbf{x;\mathcal{Z}})= \mathbf{w}_c^T (\tau_L \circ \sigma_L\circ \cdots \circ 
\tau_1 \circ \sigma_1(\mathbf{x})),
\end{equation}
where $\sigma_l: \mathbb{R}^{n_{l-1}} \to \mathbb{R}^{n_{l}}$ is the composition of the weight and activation function at the $l$-th layer of the model, $\mathbf{w}_c^T \in \mathbb{R}^{n_L}$ is the weight vector associated with class $c$, and $\tau_l(\cdot)$ applies upper bounds. 
For layer $l \in \mathcal{K}$, $\tau_l(\mathbf{v})=\min\{\mathbf{v}, \mathbf{z}_l\}$, where the ``min'' operator is applied element-wise to the vectors, ensuring that the vector values are upper-bounded by  $\mathbf{z}_l$. Otherwise, $\tau_l(\mathbf{v})=\mathbf{v}$, \textit{i.e.}, no upper bound is applied to this layer.

\textbf{Step 2: Upper Bound Optimization.}
To ensure effective defense, activation upper bounds are minimized to allow only legitimate activations through while truncating malicious ones. 
A straightforward approach is to minimize the upper bounds such that the classification accuracy of the model with these bounds, evaluated on clean samples, exceeds a certain threshold.
However, in the practical defense scenario considered here, the defender does not have access to clean, real data. 
Instead, the activation upper bounds, $\mathcal{Z}$, are optimized on the (potentially poisoned) synthetic dataset by minimizing the following Lagrangian function:
\begin{equation}\label{eq:opt_func}
    H(\mathcal{Z}, \lambda) = 
        \sum_{\mathbf{x}\in D_{\text{syn}}, c\in C}[ \overline{\mathcal{G}}_c(\mathbf{x;\mathcal{Z}}) 
        -\mathcal{G}_c(\mathbf{x})]^2 + \lambda {\sum_{l\in \mathcal{K}} {\lVert \mathbf{z}_l \rVert}_2},
\end{equation}
where $\overline{\mathcal{G}}_c(\mathbf{x;\mathcal{Z}})$ and $\mathcal{G}_c(\mathbf{x})$ represent the logits of the bounded and original global models, respectively.

We now elaborate on the optimization function:
\textbf{(i) Aligning Logits:}
Since the defender lacks access to clean, real data, optimizing for classification accuracy directly is not feasible. Moreover, the synthetic dataset may include mislabeled poisoned samples, making it ineffective to maximize classification accuracy on synthetic data.
To mitigate the impact of mislabeled synthetic samples and ensure effective defense, the first term aligns the logits of the global model with bounds, $\overline{\mathcal{G}}_c(\mathbf{x;\mathcal{Z}})$, with those of the original model, $\mathcal{G}_c(\mathbf{x})$. 
This alignment ensures the model with bounds retains similar functionality on synthetic data while avoiding reliance on potentially incorrect labels.
\textbf{(ii) Minimizing Upper Bounds:}
The second term penalizes the $l_2-$norm of the upper bounds $\mathcal{Z}$, encouraging minimal upper bounds to effectively constrain activations.
\textbf{(iii) Preserving Model Parameters:}
To maintain the model's original functionality and ensure the defense does not interfere with training, only the upper bounds $\mathcal{Z}$ are optimized, leaving the model parameters unchanged.

\textbf{Step 3: Hyper-parameter Adjustment.}
To ensure effective defense, the Lagrangian multiplier $\lambda$ is dynamically adjusted to balance global model performance and defense effectiveness.
A hard threshold to prevent a significant drop in classification accuracy on synthetic data can undermine defense effectiveness, as discrepancies between synthetic and real data may reduce accuracy and affect the defense's ability to constrain malicious activations.
To address this, we measure the relative performance on synthetic data for hyper-parameter adjustment. Specifically, a threshold $\Delta\pi$ represents the allowable drop in classification accuracy on synthetic data. A lower $\Delta\pi$ preserves performance but may fail to adequately constrain misclassifications, whereas a higher $\Delta\pi$ enforces stricter bounds but risks overly constraining the model.
The adjustment mechanism is as follows: If the accuracy drop is below $\Delta\pi$, we increase $\lambda$ by $\alpha$ to tighten the bounds; if it exceeds $\Delta\pi$, we decrease $\lambda$ to relax the constraints.

Note that after the backdoor mitigation on server, only the model parameters $\mathcal{G(\cdot)}$ are distributed to clients, excluding the upper bounds $\mathcal{Z}$. 
Separating the upper bounds from local training prevents compromised clients in classic attacks from adapting poisoned parameters to the bounds and thus evading the defense. 
The upper bounds will be distributed to the clients when the FL converges.
Furthermore, fewer than five optimization iterations are sufficient to ensure defense effectiveness, with the number of optimization parameters being significantly smaller than the entire model, ensuring efficient FL training.

\SetKwInput{Input}{Input}
\SetKwProg{kwDefense}{\textcolor{blue}{Server Backdoor Mitigation}}{}{}
\SetKwProg{kwServer}{\textcolor{blue}{Server Global Model Fusion}}{}{}
\SetKwProg{kwClient}{\textcolor{blue}{Client Update}}{}{}
\SetKwProg{init}{\textcolor{blue}{Initialization}}{}{}

\begin{algorithm}[ht]
    \small
    \SetAlgoLined
    \DontPrintSemicolon
    \Input{Lagrangian multiplier $\lambda$, threshold $\Delta\pi$, factor $\alpha$
    }
    \init{}{
        The server queries the FM to generate $\mathcal{D}_{\text{syn}}$. \\
        Randomly initialize the activation bounds $\mathcal{Z}$.
    }
    \For{\textup{each communication round} $t = 1, \cdots, T$} {
        \kwClient{}{
            Update local model $g_i$ with local dataset and upload to the server.
        }
        \kwServer{}{
            $\mathcal{G} \leftarrow \mathcal{E}(\{g_i\}, \mathcal{D}_{\text{syn}})$
            The aggregation function $\mathcal{E}(\cdot)$ is from \cite{FedDF,FedMD}.
        }
        \kwDefense{}{
            Apply bounds $\mathcal{Z}$ to $\mathcal{G}$ to obtain $\overline{\mathcal{G}}$ (Eq.\ref{eq:bounded_activation}). \\
            Update $\mathcal{Z}$ by minimizing $H(\mathcal{Z}, \lambda)$ (Eq.\ref{eq:opt_func}). \\
            \eIf{$\mathcal{G}(\mathcal{D}_{\text{syn}}) - \overline{\mathcal{G}}(\mathcal{D}_{\text{syn}}, \mathcal{Z})<\Delta\pi$}
            {
            $\lambda *= \alpha$ \\
            }
            {
            $\lambda /= \alpha$ \\
            }
        }
        Distribute the global model parameters $\mathcal{G}(\cdot)$ (without the activation bounds $\mathcal{Z}$) to clients. \\
    }
    Distribute the global model parameters with bounds $\overline{\mathcal{G}}(\cdot;\mathcal{Z})$ to clients. \\
\caption{{\small \text{Algorithm flow of the proposed defense.}}}
\label{alg:defense}
\end{algorithm}
\section{Experiment}\label{sec:exp}

\begin{table*}[ht]
\centering
\vspace{-0.1in}
\caption{Defenses against the novel backdoor attack in homo-FL. D1: CIFAR-10, D2: CIFAR-100}
\resizebox{1\textwidth}{!}{
\scriptsize
\begin{tabular}{p{.15cm}p{.85cm}|p{.3cm}p{.3cm}p{.3cm}p{.3cm}p{.3cm}p{.3cm}p{.3cm}p{.3cm}p{.3cm}p{.3cm}p{.3cm}p{.3cm}p{.3cm}p{.3cm}p{.3cm}p{.3cm}p{.3cm}p{.4cm}}
\toprule
\hline
\multicolumn{2}{c|}{\multirow{2}{*}{\textbf{Data}}} & \multicolumn{2}{c}{\textbf{Vanilla}} & \multicolumn{2}{c}{\textbf{NormThr}} & \multicolumn{2}{c}{\textbf{DP}} & \multicolumn{2}{c}{\textbf{Krum}} & \multicolumn{2}{c}{\textbf{ClipCluster}} & \multicolumn{2}{c}{\textbf{SignGuard}} & \multicolumn{2}{c}{\textbf{RFOUT}} & \multicolumn{2}{c}{\textbf{Pruning}} & \multicolumn{2}{c}{\textbf{Ours}} \\
& & ACC & ASR & ACC$\downarrow$ & ASR & ACC$\downarrow$ & ASR & ACC$\downarrow$ & ASR & ACC$\downarrow$ & ASR & ACC$\downarrow$ & ASR & ACC$\downarrow$ & ASR & ACC$\downarrow$ & ASR & ACC$\downarrow$ & ASR \\
\hline
\multicolumn{20}{c}{\textbf{Cross-Silo}} \\
\hline
\multirow{2}{*}{$\boldsymbol{D_1}$} 
& {\scriptsize\textbf{IID}} & 81.56 & 92.67 & 3.14 & 72.42 & 15.28 & 80.24 & 1.72 & 93.36 & 0.50 & 92.83 & 0.21 & 92.77 & 0.02 & 92.84 & 0.56 & 84.79 & 3.14 & \textbf{4.68} \\
& {\scriptsize\textbf{non-IID}} & 94.33 & 90.43 & 0.74  & 71.13 & 18.45 & 69.27 & 44.44 & 83.70 & 0.28 & 89.40 & 0.20 & 89.80 & 0.29 & 90.43 & 0.67 & 62.98 & 0.16 & \textbf{19.15} \\
\multirow{2}{*}{$\boldsymbol{D_2}$} 
& {\scriptsize\textbf{IID}}  & 37.90 & 88.99 & 3.46 & 70.13 & 15.90 & 67.18 & 1.14 & 87.09 & 0.08 & 89.00 & 0.00 & 88.99 & 0.12 & 88.98 & 1.22 & 77.84 & 4.70 & \textbf{8.34} \\
& {\scriptsize\textbf{non-IID}}  & 60.65 & 81.61 & 3.75 & 45.51 & 3.99 & 43.74 & 12.74 & 79.17 & 0.45 & 79.99 & 0.19 & 81.12 & 0.02 & 81.50 & 1.89 & 64.85 & 4.60 & \textbf{2.47} \\
\hline
\multicolumn{20}{c}{\textbf{Cross-Device}} \\
\hline
\multirow{2}{*}{$\boldsymbol{D_1}$} 
& {\scriptsize\textbf{IID}} & 63.92 & 96.31 & 4.41 & 95.53 & 6.41 & 96.29 & 0.30 & 96.32 & 0.12 & 96.37 & 0.02 & 96.39 & 0.24 & 96.35 & 0.56 & 84.79 & 4.08 & \textbf{11.81} \\
& {\scriptsize\textbf{non-IID}}  & 88.26 & 92.90 & 12.90 & 89.50 & 16.93 & 90.16 & 17.05 & 92.74 & 10.07 & 95.92 & 0.38 & 92.92 & 0.06 & 92.72 & 1.48 & 71.60 & 1.75 & \textbf{19.67} \\
\multirow{2}{*}{$\boldsymbol{D_2}$} 
& {\scriptsize\textbf{IID}}  & 20.30 & 90.96 & 2.55 & 82.18 & 11.40 & 82.20 & 1.30 & 89.57 & 0.52 & 90.94 & 0.36 & 90.98 & 0.44 & 90.94 & 0.70 & 83.79 & 2.66 & \textbf{13.98} \\
& {\scriptsize\textbf{non-IID}}  & 53.06 & 89.24 & 3.39 & 55.29 & 3.66 & 53.90 & 11.68 & 79.59 & 0.29 & 89.13 & 0.08 & 89.20 & 0.09 & 89.17 & 0.15 & 64.78 & 2.45 & \textbf{8.91} \\
\hline
\bottomrule
\end{tabular}
}
\label{tab:defenses_homo}
\end{table*}

\begin{table*}[ht]
\centering
\caption{Defenses against the novel backdoor attack in hete-FL. D1: CIFAR-10, D2: CIFAR-100}
\resizebox{1\textwidth}{!}{
\scriptsize
\begin{tabular}{p{.15cm}p{.85cm}|p{.3cm}p{.3cm}p{.3cm}p{.3cm}p{.3cm}p{.3cm}p{.3cm}p{.3cm}p{.3cm}p{.3cm}p{.3cm}p{.3cm}p{.3cm}p{.3cm}p{.3cm}p{.3cm}p{.3cm}p{.4cm}}
\toprule
\hline
\multicolumn{2}{c|}{\multirow{2}{*}{\textbf{Data}}} & \multicolumn{2}{c}{\textbf{Vanilla}} & \multicolumn{2}{c}{\textbf{NormThr}} & \multicolumn{2}{c}{\textbf{DP}} & \multicolumn{2}{c}{\textbf{Krum}} & \multicolumn{2}{c}{\textbf{ClipCluster}} & \multicolumn{2}{c}{\textbf{SignGuard}} & \multicolumn{2}{c}{\textbf{RFOUT}} & \multicolumn{2}{c}{\textbf{Pruning}} & \multicolumn{2}{c}{\textbf{Ours}} \\
& & ACC & ASR & ACC$\downarrow$ & ASR & ACC$\downarrow$ & ASR & ACC$\downarrow$ & ASR & ACC$\downarrow$ & ASR & ACC$\downarrow$ & ASR & ACC$\downarrow$ & ASR & ACC$\downarrow$ & ASR & ACC$\downarrow$ & ASR \\
\hline
\multicolumn{20}{c}{\textbf{Cross-Silo}} \\
\hline
\multirow{2}{*}{$\boldsymbol{D_1}$} 
& {\scriptsize\textbf{IID}} & 79.52 & 93.36 & 3.28 & 77.39 & 16.22 & 87.35 & 0.52 & 93.74 & 1.22 & 93.47 & 0.34 & 93.75 & 0.40 & 93.74 & 2.90 & 72.55 & 2.46 & \textbf{5.98} \\
& {\scriptsize\textbf{non-IID}} & 94.31 & 91.56 & 1.48 & 87.54 & 3.64 & 87.60 & 31.58 & 89.02 & 0.36 & 89.30 & 0.32 & 90.98 & 0.21 & 91.03 & 0.69 & 64.73 & 0.00 & \textbf{20.17} \\
\multirow{2}{*}{$\boldsymbol{D_2}$} 
& {\scriptsize\textbf{IID}} & 36.52 & 89.31 & 3.76 & 69.82 & 14.70 & 64.65 & 0.10 & 87.96 & 1.58 & 89.35 & 0.80 & 89.22 & 0.48 & 89.26 & 1.14 & 81.05 & 3.80 & \textbf{9.92} \\
& {\scriptsize\textbf{non-IID}} & 61.83 & 84.19 & 3.92 & 55.04 & 4.15 & 51.78 & 6.04 & 85.92 & 1.52 & 84.24 & 0.27 & 84.76 & 0.14 & 64.77 & 1.12 & 71.01 & 4.18 & \textbf{3.36} \\
\hline
\multicolumn{20}{c}{\textbf{Cross-Device}} \\
\hline
\multirow{2}{*}{$\boldsymbol{D_1}$} 
& {\scriptsize\textbf{IID}} & 63.48 & 96.44 & 4.00 & 95.55 & 7.20 & 95.95 & 5.20 & 96.40 & 4.48 & 95.37 & 0.10 & 96.46 & 0.10 & 96.38 & 1.72 & 87.19 & 2.06 & \textbf{16.57} \\
& {\scriptsize\textbf{non-IID}} & 87.17 & 92.22 & 6.78 & 88.86 & 5.23 & 89.42 & 22.84 & 91.55 & 14.73 & 95.66 & 0.38 & 92.33 & 0.34 & 92.41 & 2.21 & 72.91 & 1.41 & \textbf{23.13} \\
\multirow{2}{*}{$\boldsymbol{D_2}$} 
& {\scriptsize\textbf{IID}} & 21.24 & 90.90 & 3.70 & 80.34 & 9.60 & 81.95 & 0.75 & 89.04 & 0.10 & 90.93 & 0.02 & 90.95 & 4.24 & 92.81 & 0.74 & 84.06 & 3.38 & \textbf{14.17} \\
& {\scriptsize\textbf{non-IID}} & 52.76 & 89.02 & 3.95 & 58.92 & 4.57 & 58.96 & 8.94 & 83.28 & 0.12 & 89.16 & 0.34 & 93.00 & 0.24 & 89.12 & 0.44 & 62.22 & 1.45 & \textbf{9.69} \\
\hline
\bottomrule
\end{tabular}
}
\label{tab:defenses_hete}
\end{table*}

\subsection{Experimental Setup}\label{sec:exp_setup}
\textbf{Datasets and models:} 
We use two benchmark datasets, \textbf{CIFAR-10} and \textbf{CIFAR-100}, for image classification \cite{dataset}. CIFAR-10 has 60k 32×32 color images across 10 classes, with 5k images per class for training and 1k per class for testing, while CIFAR-100 includes the same number of images across 100 classes, with 500 images per class for training and 100 for testing. 
For the foundation models, we employ \textbf{GPT-4} to produce prompts guiding \textbf{Dall-E} to produce 10,000 synthetic data for each dataset, with an equal distribution across all classes.
For downstream models in FL systems, we use \textbf{ResNet-18} \cite{Resnet18} with added linear layers to simulate heterogeneous FL models. \\
\textbf{FL settings:} 
We consider both homogeneous (\textbf{homo-FL}) and heterogeneous (\textbf{hete-FL}) federated learning settings, along with \textbf{cross-device} and \textbf{cross-silo} scenarios. 
In the cross-device scenario, 100 clients are available, with 10\% randomly selected for each global round. In the cross-silo scenario, 10 clients participate in every round. 
In all FL settings, we consider both \textbf{IID} (independent and identically distributed) and \textbf{non-IID} local data, where non-IID is simulated using a Dirichlet distribution with $\beta$ (the parameter deciding the degree of data heterogeneity) set to 0.1. \\
\textbf{Training settings:}
FL global rounds are set to 50, with 5 iterations for local training, ensemble distillation, and bound optimization. A learning rate of $1\times10^{-3}$ is used for local training and $5\times10^{-4}$ for distillation and bound optimization. \\
\textbf{Attack settings:} 
For the novel attack mechanism, we focus on the classic backdoor attack \textbf{BadNet}\cite{2017BadNets}. 
For the classic attack strategy, we also consider \textbf{Blend}\cite{Blend}, and \textbf{SIG}\cite{SIG}.
For all datasets, class 0 is chosen as the target class, and all trigger-embedded instances are mislabeled as class 0. 
The poisoning ratio for synthetic datasets is set to 20\%, \textit{i.e.}, 20\% of instances per non-target class are embedded with triggers. \\
\textbf{Defense settings:} 
ResNet-18 consists of four stages of residual blocks, each composed of a series of convolutional layers. To ensure the effectiveness and efficiency of the defense method, we define $\mathcal{K}$ as the set of final layers from the four stages of ResNet-18.
The initial value of the Lagrangian multiplier $\lambda$ is set to 1, and $\alpha$ is set to 1.1. \\
\textbf{Evaluation metrics:} 
We define accuracy (\textbf{ACC}) as the fraction of clean test samples correctly classified, and Attack Success Rate (\textbf{ASR}) as the fraction of backdoor-triggered samples misclassified to the target class. The defense effectiveness is evaluated by (i) the average ACC of client models on their local test sets and (ii) the average ASR on trigger-embedded test sets. A lower ASR indicates better defense, while ACC should remain as close as possible to the model's original performance without defense. \\
\textbf{Performance Evaluation:} 
To demonstrate the effectiveness of our defense, we compare its performance with other FL defense methods, including \textbf{NormThr}\cite{NormThr}, \textbf{DP}\cite{DP}, \textbf{Krum}\cite{Krum}, \textbf{Clipcluster}\cite{clipcluster},  \textbf{SignGuard}\cite{signguard},  \textbf{RFOUT}\cite{rfout}, and \textbf{Pruning}\cite{Pruning}. 
For all defense methods, including ours, we adjust the hyperparameters so that the drop in ACC is within $\Delta\pi=10\%$.


\begin{table}[ht]
    \centering
    \caption{Proposed defense against classic backdoor attacks on CIFAR-10 in cross-silo homo-FL.}
    \footnotesize
    \begin{tabular}{c|p{.4cm}p{.4cm}p{.4cm}p{.5cm}|p{.4cm}p{.4cm}p{.4cm}p{.5cm}}
    \toprule
    \hline
    \multirow{2}{*}{\textbf{Attack}} & \multicolumn{4}{c|}{\textbf{Vanilla}} & \multicolumn{4}{c}{\textbf{Ours}} \\
    & \multicolumn{2}{c}{\textbf{IID}} & \multicolumn{2}{c|}{\textbf{non-IID}} & \multicolumn{2}{c}{\textbf{IID}} & \multicolumn{2}{c}{\textbf{non-IID}} \\
    & ACC & ASR & ACC & ASR & ACC$\downarrow$ & ASR & ACC$\downarrow$ & ASR \\
    \hline
    \textbf{No Attack} & 81.36 & - & 94.53 & - & 1.48 & - & 0.24 & - \\
    \textbf{BadNet} & 82.54 & 96.50 & 94.57 & 60.38 & 3.50 & 29.25 & 0.69 & 30.08  \\
    \textbf{Blend} & 82.04 & 90.04 & 95.07 & 92.36 & 6.80 & 41.27 & 1.43 & 37.36  \\
    \textbf{SIG} & 81.46 & 85.42 & 93.97 & 73.52 & 6.36 & 22.43 & 2.14 & 36.34 \\
    \hline
    \bottomrule
    \end{tabular}
    \label{tab:defense_BD_FL}
\end{table}

\vspace{-.1in}
\subsection{Experimental Results}

\noindent\textbf{Against Novel Backdoor Attack.}
The vanilla FM-FL system demonstrates a high susceptibility to the novel backdoor attack across all FL settings and datasets. 
As shown under ``Vanilla'' in Table~\ref{tab:defenses_homo} and Table~\ref{tab:defenses_hete}, the ASRs exceed 90\% in all configurations, while the ACC on clean samples remains at reasonable levels, highlighting the stealthy nature and effectiveness of this attack.

Existing defense methods, including \textbf{NormThr}, \textbf{DP}, \textbf{Krum}, \textbf{ClipCluster}, \textbf{SignGuard}, \textbf{RFOUT}, and \textbf{Pruning}, exhibit limited effectiveness against the novel threat. 
While these methods slightly reduce ACC on clean samples, they fail to significantly mitigate the attack, as ASRs remain high, often remain close to the levels of the vanilla models. 
This verifies that existing approaches, which rely heavily on identifying anomalies in client updates, are ill-suited to counter novel backdoor attacks that stem from poisoned synthetic data on the server side.

In contrast, our proposed method offers a robust defense against the novel backdoor attack. 
For homo-FL settings (Table~\ref{tab:defenses_homo}), our defense achieves a minimal drop in ACC (ACC$\downarrow$ of less than 5\% across all datasets), ensuring that the global model retains its functionality on clean samples. 
Furthermore, our method reduces ASRs to below 15\% for IID datasets and below 20\% for non-IID datasets, demonstrating effective mitigation of the attack. 
Notably, neither the number of clients nor the level of data heterogeneity significantly affects the performance of our defense, suggesting its robustness across varying FL scenarios.

\noindent\textbf{Against Classic Backdoor Attack.}
Table~\ref{tab:defense_BD_FL} demonstrates the effectiveness of our defense method against classic backdoor attacks, including BadNet, Blend, and SIG, in the homo-FL cross-silo setting on CIFAR-10. 
To ensure the attacks remain potent, we compromise two clients and adjust the attack hyperparameters, resulting in most ASRs above 85\% for vanilla models. 
In most cases, our method reduces the ASR by more than 50\%, while keeping the ACC drop below 10\%. 
For BadNet, Blend, and SIG, ASR decreases substantially in IID settings with minimal impact on ACC. In non-IID settings, ASR drops to approximately 35\%, similarly maintaining minimal impact on ACC.
These results demonstrate robustness across various attack strategies and FL settings, highlighting its applicability in defending against a wide range of backdoor threats in FL systems. 

\noindent\textbf{Applied in non-attack setting.}
Since our defense method does not rely on the assumption of an attack, we evaluate its performance in clean (attack-free) settings to ensure it does not degrade model accuracy unnecessarily. 
As shown under ``No Attack'' in Table~\ref{tab:defense_BD_FL}, the drop in ACC remains within 2\% in the homogeneous cross-silo FL setting. 
This minimal impact demonstrates the effectiveness of our defense in maintaining model performance on clean data. Furthermore, the negligible drop in ACC highlights the method's practicality for deployment in real-world FL systems, where attacks may not always occur but defenses must remain active to ensure security without compromising model functionality.

\section{Conclusion} \label{sec:conclusion}


This paper introduces the first data-free defense strategy to address emerging backdoor threats resulting from the integration of FMs into FL. 
Within the proposed defense framework, novel and classic FL backdoor attacks are unified.
Extensive experiments conducted across diverse FL scenarios validate the effectiveness of our defense method, demonstrating its robustness and applicability in various contexts.

\bibliographystyle{plain}
\bibliography{ref}

\end{document}